\documentclass{article}

\usepackage[preprint]{neurips_2026}

\usepackage[utf8]{inputenc} 
\usepackage[T1]{fontenc}    
\usepackage{hyperref}       
\usepackage{url}            
\usepackage{booktabs}       
\usepackage{amsfonts}       
\usepackage{nicefrac}       
\usepackage{microtype}      
\usepackage{xcolor}         
\usepackage{graphicx}
\usepackage{placeins}

\title{Representation Without Control: \\Testing the Realization Effect in Language Models}

\author{%
  Ciarán Walsh \\ 
  Columbia University\\
  \texttt{crw2161@columbia.edu} \\
  \And
  Emilio Barkett \\
  Columbia University \\
  \texttt{eab2291@columbia.edu} \\
}

\begin{document}

\maketitle

\begin{abstract}
Large language models are increasingly used as behavioral simulators, but it remains unclear when their outputs reflect human-like cognitive mechanisms rather than prompt-sensitive surface patterns. We study this question through the realization effect, a well-characterized finding in behavioral economics in which risk-taking differs systematically after paper versus realized gains and losses. We evaluate LLM behavior at three levels: prompt-only behavioral sensitivity, linear readout of internal representations, and causal control via activation steering. Prompt-only results show systematic condition sensitivity, but the directional pattern does not reproduce human realization-effect predictions. Gemma's residual stream contains a linearly decodable realization-status signal at layer 18 that generalizes to held-out prompts. Steering along this direction does not, however, reliably shift downstream risk choices, a null result that holds across positive scales and in a negative sign-symmetry run. Behavioral sensitivity, latent readout, and causal control are three distinct properties that do not automatically co-occur, and successful latent readout is insufficient evidence that a model behaviorally relies on a representation during downstream decision-making.
\end{abstract}

\section{Introduction}

A growing body of research treats large language model outputs as proxies for human cognition, but condition-sensitive behavior and genuine cognitive implementation are not the same thing. This practice rests on an implicit assumption: that when a model produces psychologically interpretable outputs, it does so by implementing something like the human mechanism that originally motivated the task. That assumption is rarely tested directly. A model could be condition-sensitive, producing different responses under different framings, without encoding the underlying construct that makes those framings meaningful to humans. Distinguishing these two possibilities requires going beyond behavioral measurement and asking whether the relevant structure exists inside the model at all.

We study this problem through the \emph{realization effect}, a well-documented phenomenon in human decision-making whereby risk attitudes depend on whether prior gains or losses remain open (paper) or have been closed (realized). The realization effect is a useful test case because it is behaviorally well-characterized with directional predictions that are specific and non-obvious: the same dollar outcome carries different implications for subsequent risk-taking depending solely on whether the account containing it is open or closed~\cite{imas2016realization, flepp2021paper, merkle2021closing}. Realization status is also a discrete, manipulable property of a prompt, which makes it tractable to study both behaviorally and representationally.

Our investigation proceeded in two stages. We first asked whether LLMs reproduce the realization effect at the behavioral level, that is, whether prompt-only responses show the same pattern of condition-sensitivity that human subjects exhibit. Finding that behavioral replication was ambiguous, we turned to the mechanistic question: is realization status encoded as a readable internal representation, and if so, does that representation causally govern downstream risk choices? This pivot reflects a broader point about LLM behavioral research: behavioral sensitivity and mechanistic implementation are not the same thing, and conflating them can lead to misleading conclusions about what models actually do.

We test three linked claims. First, prompt-only LLM responses are sensitive to paper-versus-realized framing, but do not robustly match human realization-effect predictions. Second, realization status is linearly decodable from Gemma's residual stream, and a direction trained on one set of prompts generalizes to held-out readout prompts. Third, and most consequentially, steering along that decoded realization direction does not reliably shift downstream risk behavior. Taken together, these results carry a pointed implication for mechanistic interpretability: causal validation, not just correlational decoding, is necessary to establish that a representation is doing real work.

\section{Related Work}

\subsection{The realization effect}

The realization effect rests on a distinction between paper and realized outcomes rooted in mental accounting theory. Within an open mental account, prior gains and losses are evaluated jointly with subsequent prospects; closing the account internalizes the outcome and resets the reference point. Imas~\cite{imas2016realization} formalized this distinction to reconcile contradictory findings: prior work had found both that losses increase subsequent risk-taking and that they decrease it. The divergence tracks realization status. When losses remained unrealized, individuals chased them by taking on more risk; when losses were settled by a transfer of money between accounts, individuals took on less. The mechanism follows from cumulative prospect theory and narrow bracketing: a paper loss is integrated with subsequent prospects, making a break-even gamble especially attractive, whereas a realized loss closes the bracket and removes the possibility of recovery. Flepp et al.~\cite{flepp2021paper} extended this evidence to a field setting using casino gambling records, and Merkle et al.~\cite{merkle2021closing} replicated the effect for gains and identified positive lottery skewness as a necessary boundary condition. Together, this literature establishes directional predictions specific enough to serve as a test case for LLM behavioral evaluation: the same dollar outcome carries different implications for subsequent risk-taking depending solely on whether the account containing it is open or closed.

\subsection{LLMs as behavioral subjects}

Prior work has proposed using LLMs as simulated human or economic subjects~\cite{aher2023simulate,horton2023simulated}, and a growing literature has evaluated how well this works in practice. Such experiments are attractive because they are cheap, reproducible, and easy to scale, but replication success rates vary substantially across research domains, effect sizes, and question types~\cite{yeykelis2024using, ashokkumar2024predicting}. Model responses are sensitive to prompt wording and surface framing in ways that human responses are not~\cite{salinas2024butterfly}, and LLMs tend toward response homogeneity across personas, underrepresenting the variance that characterizes real human samples~\cite{bisbee2024synthetic}. More fundamentally, even when LLMs reproduce a behavioral pattern, the mechanism driving the response may bear no resemblance to the human mechanism that motivated the task~\cite{harding2024ai, dillion2023can}. The realization effect is particularly vulnerable to this confound: because realization status is communicated entirely through framing, a model sensitive to surface features could appear to implement the effect without encoding the underlying mental-account construct. Behavioral prompting alone is therefore weak evidence that a model represents or reasons with realization status as a latent variable.

\subsection{Activation readout and steering}

Mechanistic interpretability offers a way to move beyond prompt-level behavior, and has emerged in part as a response to the recognized limitations of behavioral evaluation alone. Linear probes and contrastive activation directions can identify variables represented in a model's hidden states, and prior work has shown that emotions, sentiment, and truthfulness-related signals are often linearly decodable from residual-stream activations~\cite{zou2023representation}. Activation steering methods then test whether adding such a direction during inference changes model behavior~\cite{turner2023activation}, providing a stronger causal test than readout alone. However, successful readout does not imply behavioral control. Park et al.~\cite{park2024linear} show that a direction can encode information about a variable without being the policy-relevant signal that governs downstream choices: representations that are linearly readable may be epiphenomenal, present in the model's activations but not causally connected to its outputs. The combination of readout and steering is therefore a strictly stronger test than either alone, and the gap between them is precisely what our experiment is designed to probe.

\subsection{This paper}

We use the realization effect as a theoretically grounded test case for LLM behavioral evaluation, apply activation readout to ask whether realization status is internally represented, and use steering to test whether that representation is causally upstream of risk behavior. Recent work has argued that behavioral sensitivity alone is insufficient grounds for treating LLMs as genuine human surrogates~\cite{gao2025caution}; our experiment is designed to test whether the stronger mechanistic standard is met.

\section{Methods}

The experimental design proceeds in three stages. First, we test whether prompt-only LLM responses show sensitivity to realization-status framing. Second, we ask whether realization status is linearly decodable from model activations, and whether a decoded direction generalizes to held-out prompts. Third, we test whether steering along that direction causally shifts downstream risk behavior. Each stage corresponds to a subsection below.

\subsection{Prompt-only behavioral task}

We constructed vignette prompts modeled on the realization-effect literature; code and data available at \href{https://github.com/EmilioBarkett/realization-effect-project}{github.com/EmilioBarkett/realization-effect-project}. The prompt set uses 11 conditions spanning paper and realized outcomes, with both absolute-outcome and balance-relative prompt versions. For each prompt, the model is asked to return two integers: a next-session wager from 1 to 1000 CHF and a slot-machine risk preference from 1 to 5. The cleaned dataset contains 54,450 rows, of which 53,547 have valid parsed wagers and 49,351 have valid risk-profile responses. Two incomplete cells were excluded: \texttt{qwen/qwq-32b} and \texttt{z-ai/glm-5} at temperature 1.5.

We analyze two outcomes: log wager and risk profile. The main regressions use condition indicators with model, temperature, and prompt-version fixed effects, and HC3 robust standard errors. Paper conditions are compared against a paper-even baseline; realized conditions are compared against a small realized-loss baseline.

\subsection{Activation-vector prompt set}

For the mechanistic pass, we generated paired prompts contrasting \texttt{paper\_open} and \texttt{realized\_closed} framings across several realization-status domains, including finance, reimbursement, budget, compensation, academic, project-outcome, and casino scenarios. The behavior-evaluation subset is narrower, containing only casino and finance prompts that ask for downstream wager or investment-risk choices.

We logged Gemma 3 4B residual-stream activations and constructed a mean-difference direction:
\[
v_{\mathrm{realization}} = \mu_{\mathrm{realized\_closed}} - \mu_{\mathrm{paper\_open}}.
\]

Throughout the paper, we distinguish two versions of this direction. The train-only layer-18 direction is computed using only the \texttt{direction\_train} split of 756 matched paper/realized pairs at layer 18; this is the direction used for all readout generalization tests and steering interventions reported as main results. The all-pairs direction uses the full activation run and appears only in descriptive plots, which are labeled accordingly. We evaluated the frozen train-only layer-18 direction on the original \texttt{direction\_val} and \texttt{behavior\_eval} splits, plus a held-out prompt set authored by DeepSeek containing 40 matched pairs divided into \texttt{heldout\_readout} and \texttt{heldout\_behavior\_eval} subsets. Neutral outcome cells were omitted from the DeepSeek set because preliminary generation produced ambiguous wording.

\subsection{Steering intervention}

We implemented a residual-stream steering runner for local Gemma 3 4B. During generation, a forward hook adds a scaled, normalized realization direction at a selected layer. All main steering results use the train-only layer-18 direction, applied at the final active token position across scales of $-50$, $0$, $+50$, $+75$, $+100$, and $+150$. Positive scales point toward \texttt{realized\_closed}; negative scales point toward \texttt{paper\_open}. Each scale is applied to the same 648 behavior-evaluation prompts, and steered responses are compared to the in-run unsteered scale-0 baseline on matched prompt IDs.

We report results for both all matched valid rows and an exactly-two-integer subset in which the response contains exactly two integer tokens. This subset matters because steering can perturb the response format, and parsing failures on malformed outputs can produce misleading numeric results.

\subsection{Positive-control classification steering}

As a diagnostic check on whether the realization direction has any causal purchase on model behavior, we tested whether steering can move a direct semantic judgment even when wager control is weak. We constructed a paired classification task from the same behavior-evaluation prompts: instead of returning wager and risk integers, the model is asked to classify each scenario as \texttt{REALIZED} or \texttt{PAPER}. Because free generation often appends extra text, we scored label candidates directly from token log-probabilities under the steering hooks and selected the higher-scoring label. Scores are length-normalized and prior-calibrated at each scale, then summarized by accuracy and realized-prediction rate.

This classification run uses the all-pairs layer-18 direction rather than the train-only layer-18 direction, and evaluates three scales ($-100$, $0$, $+100$) on all 648 behavior-evaluation prompts. One prompt-construction issue applies: the classification instruction was present in the prebuilt prompt CSV and was then appended again by the steering runner, producing doubled instructions for some prompts. Results from this subsection should therefore be treated as diagnostic evidence rather than primary findings.

\section{Results}

The results follow the three-stage structure of the methods. We first ask whether prompt-only behavior is sensitive to realization-status framing. We then ask whether realization status is linearly readable from model activations and whether a decoded direction generalizes to held-out prompts. Finally, we ask whether steering along that direction causally shifts downstream risk behavior. The answer is mixed: readout succeeds where behavioral replication and steering both fall short.

\begin{figure}[h]
  \centering
  \includegraphics[width=\textwidth]{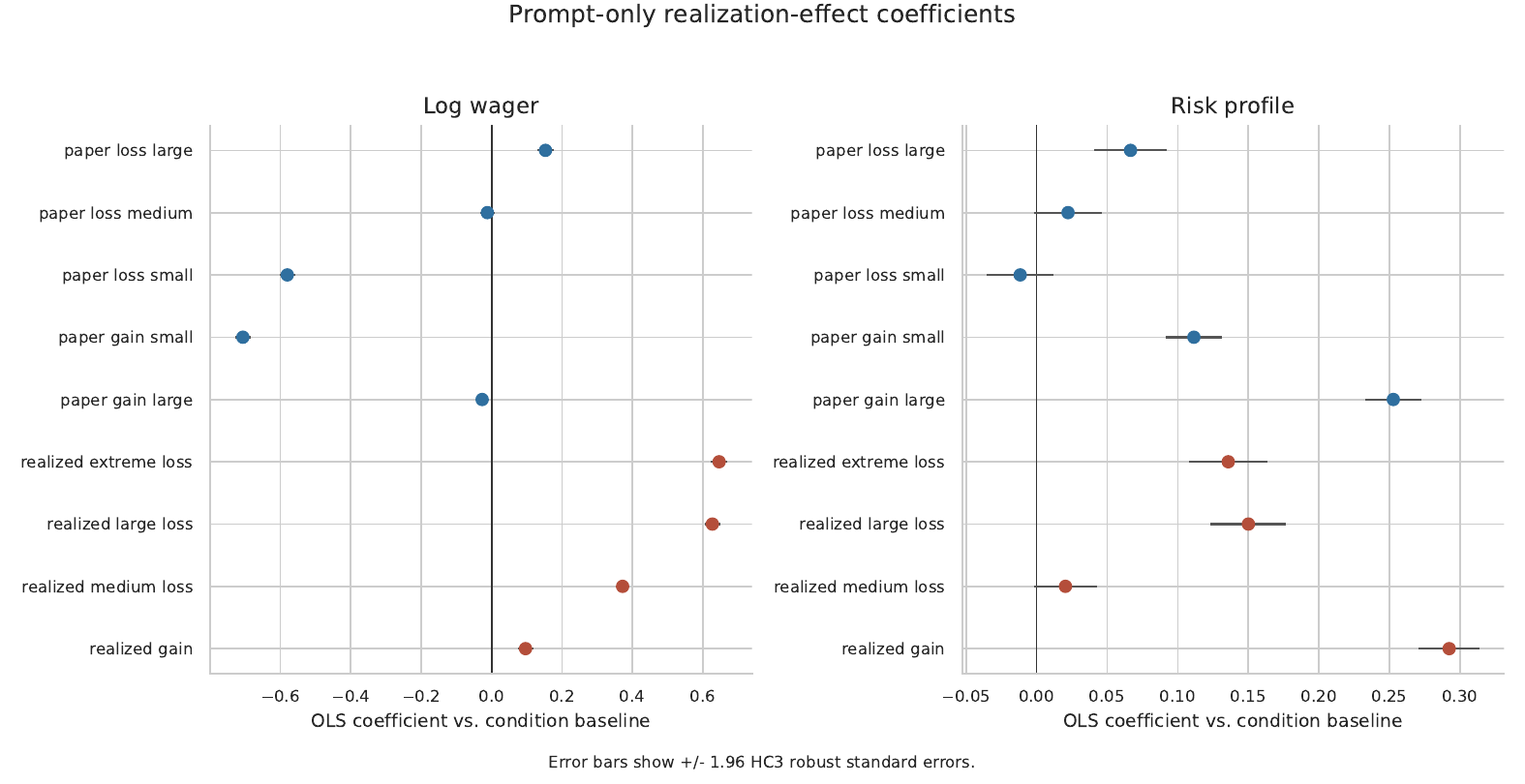}
  \caption{Prompt-only behavioral coefficients from the cleaned realization-effect dataset. Points show OLS condition coefficients; error bars show approximate 95\% intervals using HC3 robust standard errors. The model is condition-sensitive, but the coefficient pattern does not cleanly match the original human realization-effect predictions.}
  \label{fig:behavioral-coefficients}
\end{figure}

\subsection{Prompt-only behavior is condition-sensitive but not a clean realization-effect replication}

The prompt-only behavioral data show systematic responses to outcome framing, but not in the pattern predicted by the human realization-effect literature. For \texttt{risk\_profile}, paper-outcome effects are mostly positive relative to the paper baseline: paper large losses, paper small gains, and paper large gains all increase the risk-profile measure. Paper medium losses are positive but fall short of significance at the 0.05 level, and paper small losses do not significantly differ from baseline. Wager effects are more mixed: paper large losses increase log wager, while paper small losses and paper gains decrease it. Paper medium losses do not reliably differ from the paper-even baseline.

Realized-outcome results diverge more sharply from the human prediction. Realized losses increase rather than decrease log wager relative to the small realized-loss baseline, the opposite of what Imas~\cite{imas2016realization} and Flepp et al.~\cite{flepp2021paper} document in humans. For risk profile, extreme and large realized losses increase responses, while medium realized losses are not significant. Realized gains also differ significantly from baseline. The model is clearly sensitive to outcome magnitude and framing, but that sensitivity does not map cleanly onto the realization effect as documented in humans. Figure~\ref{fig:behavioral-coefficients} shows the full coefficient pattern.

\subsection{Gemma has a readable realization representation}

The activation analysis gives the positive mechanistic result. Along the train-only layer-18 direction, \texttt{paper\_open} and \texttt{realized\_closed} prompts separate clearly across the logged activation set, including the behavior-evaluation prompts used later for steering. A descriptive projection plot using the all-pairs direction is provided in Appendix~\ref{app:projection} for reference.

The stricter test is the train-only readout analysis. A direction built only from the 756 \texttt{direction\_train} pairs successfully separates realized/closed from paper/open prompts on the original held-out validation split and on the newly generated DeepSeek held-out set. Separation is strongest for direction-style prompts and weaker for behavior-evaluation prompts, particularly in the small DeepSeek behavior subset. Across all evaluated splits, realized-minus-paper mean projection deltas are positive, supporting the claim that realization status is linearly encoded in Gemma's residual stream at layer 18. Full numerical results by split are reported in Appendix~\ref{app:readout}.

\subsection{Projection strength is weakly linked to behavior}

We next asked whether stronger projection on the realization direction predicts the model's actual wager or risk response. At the raw prompt level, projection is positively associated with wager: a one-standard-deviation increase in projection predicts an 84.44 CHF larger wager ($p < 10^{-5}$). Projection also weakly predicts risk profile ($+0.136$ on the 1--5 scale, $p = 0.021$). After controlling for pair role, domain, outcome valence, amount bucket, and prompt source, however, the wager coefficient falls to $+5.21$ CHF and is no longer significant ($p = 0.776$). The risk-profile coefficient remains marginal ($+0.134$, $p = 0.053$), but the model explains little variance. The raw association reflects prompt structure rather than a genuine link between representation strength and behavior.

\begin{figure}[htbp]
  \centering
  \includegraphics[width=\textwidth]{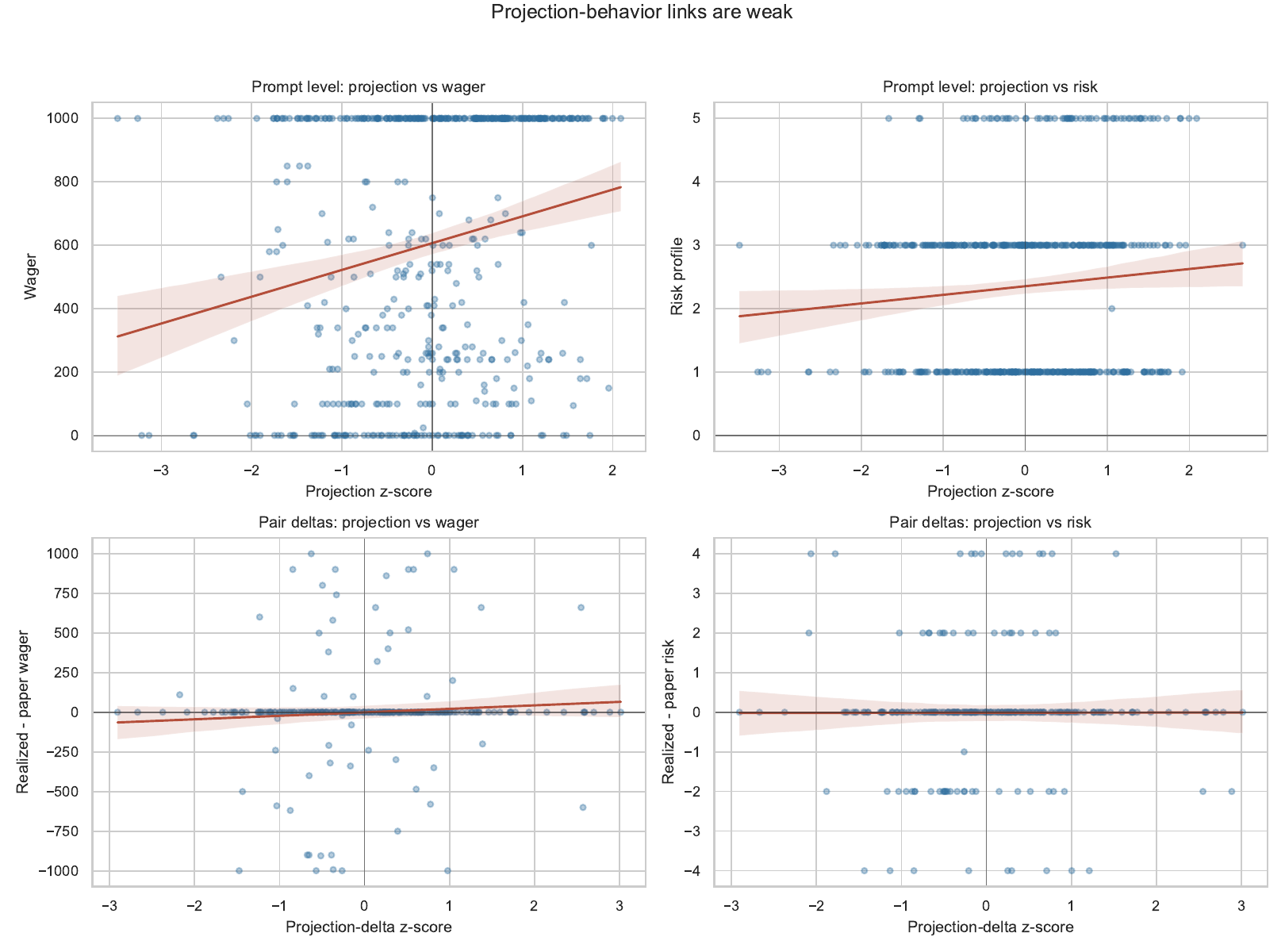}
  \caption{Projection strength versus behavior. Raw prompt-level projection has a positive association with wager, but within matched paper/realized pairs, projection deltas weakly predict wager deltas and essentially do not predict risk deltas. This supports the distinction between readable representation and behavioral control.}
  \label{fig:projection-behavior}
\end{figure}

The within-pair analysis is the more demanding test. For each matched pair, we computed the realized-minus-paper projection delta and compared it with the realized-minus-paper wager and risk deltas. Projection deltas do not predict behavior deltas: Pearson $r = 0.072$ for wager and $r = 0.001$ for risk. Regressions with prompt controls find no reliable effects. The activation direction tracks realization status, but variation in projection strength does not explain variation in downstream risk behavior. Figure~\ref{fig:projection-behavior} shows both the raw prompt-level association and the within-pair null.

\subsection{Steering does not robustly control risk behavior}

If the train-only layer-18 direction were a causal lever for risk-taking, steering toward \texttt{realized\_closed} should move behavior in one direction and steering toward \texttt{paper\_open} should move it in the opposite direction. We do not observe this pattern. Across positive scales $+50$, $+75$, $+100$, and $+150$, mean wager deltas are small and positive, but median deltas are zero throughout. Mean risk-profile deltas remain close to zero, with small negative movement at larger positive scales. Table~\ref{tab:steering-deltas} reports the full matched delta breakdown by scale.

\begin{table}[h!]
  \centering
  \caption{Matched steering deltas relative to the in-run unsteered Gemma behavior baseline. The table reports all valid matched rows from the train-only layer-18 steering run. Exactly-two-integer rows show the same qualitative result: small mean shifts and zero median shifts.}
  \label{tab:steering-deltas}
  \begin{tabular}{rrrrr}
    \toprule
    Scale & Matched rows & Mean wager delta & Mean risk delta & Median deltas \\
    \midrule
    -50 & 478 & 11.80 & 0.094 & 0 / 0 \\
     50 & 476 & 7.24 & 0.013 & 0 / 0 \\
     75 & 483 & 8.21 & -0.025 & 0 / 0 \\
    100 & 478 & 15.24 & -0.050 & 0 / 0 \\
    150 & 473 & 12.80 & -0.080 & 0 / 0 \\
    \bottomrule
  \end{tabular}
\end{table}

\begin{figure}[h!]
  \centering
  \includegraphics[width=\textwidth]{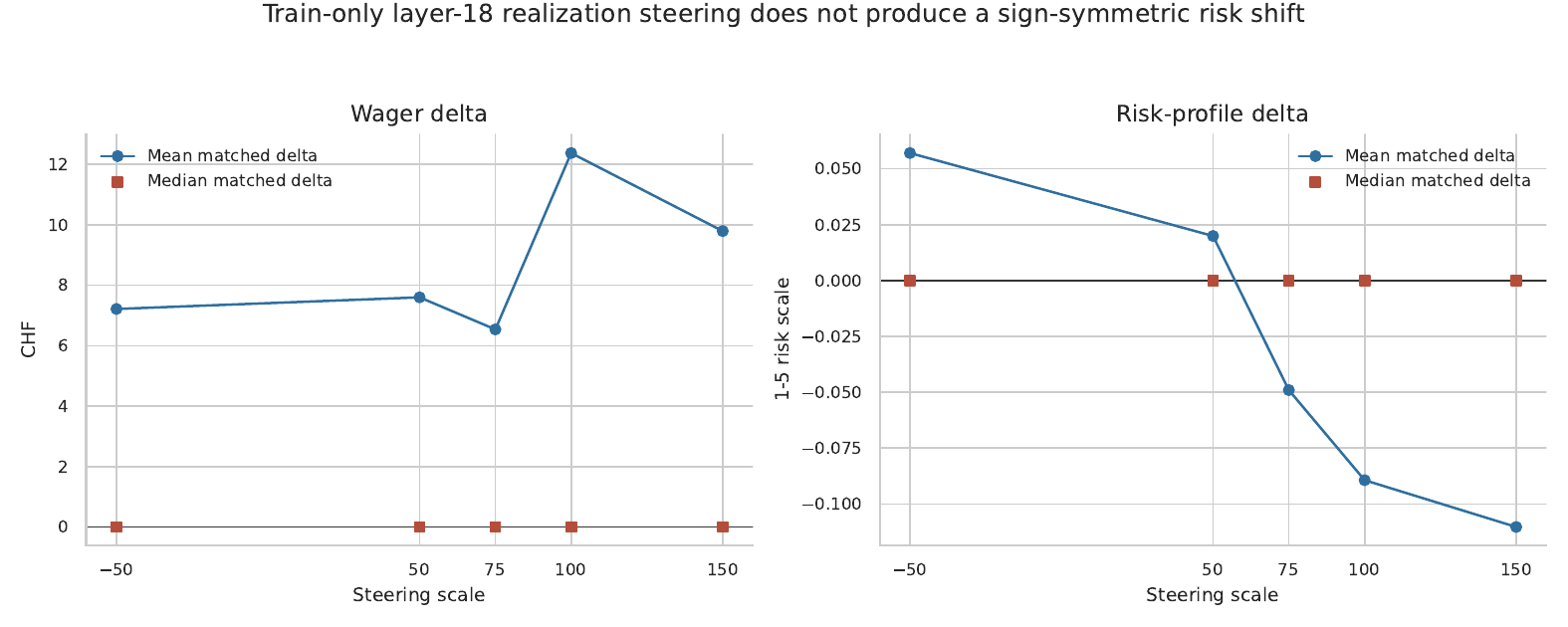}
  \caption{Dose-response plot for Gemma train-only layer-18 realization steering on the exactly-two-integer matched subset. The negative scale does not produce a clean behavioral reversal, and median matched deltas remain zero across wager and risk outcomes.}
  \label{fig:steering-dose}
\end{figure}

The negative-scale run at $-50$ also fails to produce a clean reversal: the exactly-two-integer matched mean wager delta is $+7.22$, the mean risk delta is $+0.057$, and median deltas are zero for both outcomes. The absence of sign-symmetric behavioral responses at opposite steering scales is the clearest evidence that the train-only layer-18 direction does not function as a reliable control variable for downstream risk behavior. Figure~\ref{fig:steering-dose} shows the dose-response across all evaluated scales.

\subsection{Positive-control classification steering shows only weak directional movement}

As a diagnostic check, we tested whether steering can move a direct semantic judgment even when wager control is weak. The realization direction does induce a small directional shift in the classification task: the realized-prediction rate increases from 0.140 at unsteered scale 0 to 0.198 at $+100$, and decreases to 0.120 at $-100$. Classification accuracy, however, remains near chance overall (0.520 at scale 0, 0.515 at $+100$, 0.512 at $-100$). The model strongly favors the \texttt{PAPER} label across all scales: \texttt{paper\_open} prompts are classified correctly at a rate of 0.863, while \texttt{realized\_closed} prompts are classified correctly at only 0.169. The directional shift exists but is too small and asymmetric to constitute usable behavioral control. These results should additionally be treated with caution given the duplicated-instruction issue noted in the methods.

\subsection{Compliance artifacts affect interpretation of steering results}

A methodological caution bears directly on how to interpret the steering results. The task asks for exactly two integers, and steering perturbs response format in a non-trivial fraction of cases, producing outputs that either omit one integer or add extra tokens that prevent unambiguous parsing. Noncompliance is concentrated in prompts generated from the Sonnet source, which fail the exactly-two-integer criterion at substantially higher rates than GPT-5.4-derived or Grok-fast-derived prompts. At scale $-50$, for example, Sonnet-derived prompts produce 106 noncompliant responses out of 216, compared with 46/216 for GPT-5.4-derived prompts and 70/216 for Grok-fast-derived prompts. Because parsed numeric outcomes are only meaningful when the answer contract is followed, the exactly-two-integer subset analyses reported above are the more reliable basis for interpreting steering effects. Full compliance figures by scale and prompt source are provided in Appendix~\ref{app:compliance}.

\begin{table}[h!]
  \centering
  \small
  \caption{Hypothesis outcome summary.}
  \label{tab:hypothesis-summary}
  \begin{tabular}{p{0.26\textwidth}p{0.15\textwidth}p{0.49\textwidth}}
    \toprule
    Test & Outcome status & Evidence summary \\
    \midrule
    Prompt-only behavioral replication & Weak support & Condition sensitivity is clear, but the directional pattern does not robustly match human realization-effect predictions. \\
    Activation readout (layer-18 direction) & Supported & A train-only direction separates realized/closed from paper/open prompts on original validation prompts and on a newly generated DeepSeek held-out readout set. \\
    Projection strength versus behavior & Not supported & Within-pair projection deltas do not predict wager or risk deltas after controlling for prompt structure. \\
    Risk-behavior steering intervention & Not supported & Steering shifts are small, medians are zero, and negative-scale symmetry does not produce a clean reversal. \\
    Positive-control classification steering & Weak diagnostic support & Directional shift in realized-prediction rate occurs across scales, but accuracy remains near chance with a strong \texttt{PAPER} prediction bias and a duplicated-instruction caveat. \\
    \bottomrule
  \end{tabular}
\end{table}

\FloatBarrier

\section{Discussion}

The central result is a dissociation between linear readout and causal control. Gemma contains a linearly decodable realization-status signal that generalizes beyond the prompts used to build the train-only layer-18 direction, but the corresponding activation direction does not reliably control downstream risk choices. This dissociation has implications for both behavioral evaluation and mechanistic interpretability.

For behavioral evaluation, the prompt-only results warn against interpreting LLM condition sensitivity as evidence of human-like cognition. The models react to prior outcomes and are sensitive to framing, but the response pattern does not reproduce the mental-accounting signature documented in humans. This is consistent with the broader concern that LLMs may pass behavioral tests by tracking surface features of prompts rather than implementing anything like the underlying decision mechanism~\cite{gao2025caution, harding2024ai}. The realization effect illustrates this risk cleanly: because realization status is communicated entirely through framing, a model sensitive to prompt wording could appear to implement the effect without encoding the construct at all.

For mechanistic interpretability, the steering results warn against treating linear decodability as explanation. A representation can be readable in residual-stream activations while failing to function as a causal control variable~\cite{park2024linear}. Our experiment gives this theoretical concern concrete empirical content: readout succeeds where steering fails, precisely the dissociation that motivates the causal turn in interpretability research. Of the possible explanations for the steering failure, the policy-routing account is most plausible. The train-only layer-18 direction may track semantic status without connecting to the numeric decision policy governing wager and risk outputs, which are likely dominated by instruction-following priors, learned defaults, or numeric anchors more strongly weighted than the realization representation at the point of output. The sign-symmetry result argues against a simple magnitude explanation: if insufficient steering strength were the problem, negative scales should produce at least a directional reversal, and they do not.

The results suggest that behavioral sensitivity, latent readout, and causal control are three distinct evidential standards that do not automatically co-occur, and that the gap between the second and third is non-trivial and empirically detectable. Whether the dissociation is specific to numeric output tasks, to the realization construct, or to Gemma remains to be determined, but the finding motivates treating causal control as a necessary condition for mechanistic claims rather than an optional extension.

\section{Limitations}

The intervention reported here is narrow by design: a single model, a single layer, a normalized mean-difference direction, and final-token steering. This makes the results interpretable but limits their scope. All steering findings are specific to Gemma 3 4B at layer 18, and we have not tested whether a layer sweep, a position-mode sweep, or multi-layer steering would produce stronger behavioral effects. Hosted models such as Qwen support prompt-only replication but not residual-stream steering, since hidden states and generation hooks are not exposed through API access. The null result should therefore be read as specific to this configuration rather than as evidence that no steering approach could move risk behavior.

The behavioral assay introduces its own constraints. Wager and risk-profile answers are constrained numeric outputs, which makes scoring tractable but may amplify format and parsing artifacts. Exactly-two-integer compliance is imperfect, particularly for Sonnet-derived prompts, and the prompt-only dataset pools 25 models, so aggregate coefficients should be interpreted alongside per-model robustness checks. The decoded direction may also track a broad semantic contrast between open and closed framings rather than the specific decision-relevant variable that would be needed to shift risk-taking, a possibility the current design cannot rule out.

On held-out evidence, the DeepSeek set is useful because it was not used to construct the direction and passed a local overlap audit, but it is small and single-source. The \texttt{heldout\_behavior\_eval} split tests projection separation only, not newly generated downstream wager or risk responses, which limits what can be concluded about generalization of the behavioral link. Finally, the steering summaries report matched mean and median deltas without bootstrap or randomization intervals. The zero medians and absence of sign-symmetric responses are the primary evidence against robust control, but calibrated interval estimates would make the null result more precisely quantified.

\section{Future Work}

The most important open question is whether the steering null result is specific to the layer-18 final-token configuration or reflects a deeper limit. Layer sweeps across layers 14 through 22, combined with \texttt{position\_mode=all} and multi-layer steering schedules, would distinguish between these explanations. The positive-control classification run should also be rerun with the train-only layer-18 direction and corrected prompt construction before being treated as more than diagnostic.

Expanding held-out evaluation is a second priority. Adding multiple prompt-source models, larger behavior-evaluation subsets, and newly generated Gemma responses for held-out behavior prompts would allow the representation and behavioral results to be evaluated independently, and would clarify whether the readout-behavior gap is a stable feature of the model or an artifact of the current prompt set.

The most consequential extension would be cross-model replication. If the same dissociation between readable representation and failed behavioral control appears in Qwen or other architectures under the same local hook framework, it would shift the interpretation from a Gemma-specific artifact to a structural feature of how language models route semantic representations into numeric outputs, considerably strengthening the generalizability of the main finding.

\section{Conclusion}

We asked whether LLMs reproduce the realization effect in risk-taking and whether behavioral sensitivity reflects genuine internal implementation of the underlying construct. The answers are no and not demonstrably. Prompt-only behavior is condition-sensitive but does not match human realization-effect predictions. Gemma encodes a linearly decodable realization-status signal at layer 18 that generalizes to held-out prompts, but projection strength weakly predicts behavior within matched pairs and steering along the train-only layer-18 direction does not reliably shift risk choices. The negative result is the contribution: behavioral sensitivity, latent readout, and causal control are three distinct properties that do not automatically co-occur, and a model can pass the first two tests while failing the third. Condition sensitivity is weak evidence of mechanism and linear decodability is weak evidence of causal relevance. Establishing that a model genuinely implements a human decision-making construct requires showing that the relevant representation is causally upstream of the behavior of interest, a harder bar than the field currently applies and one that should be the standard.

\begin{ack}
The authors thank the Supervised Program for Alignment Research (SPAR) for supporting this project.
\end{ack}

\bibliographystyle{plain}
\bibliography{references.bib}

\appendix

\newpage

\section{Descriptive Activation Projection}
\label{app:projection}

\begin{figure}[h!]
  \centering
  \includegraphics[width=\textwidth]{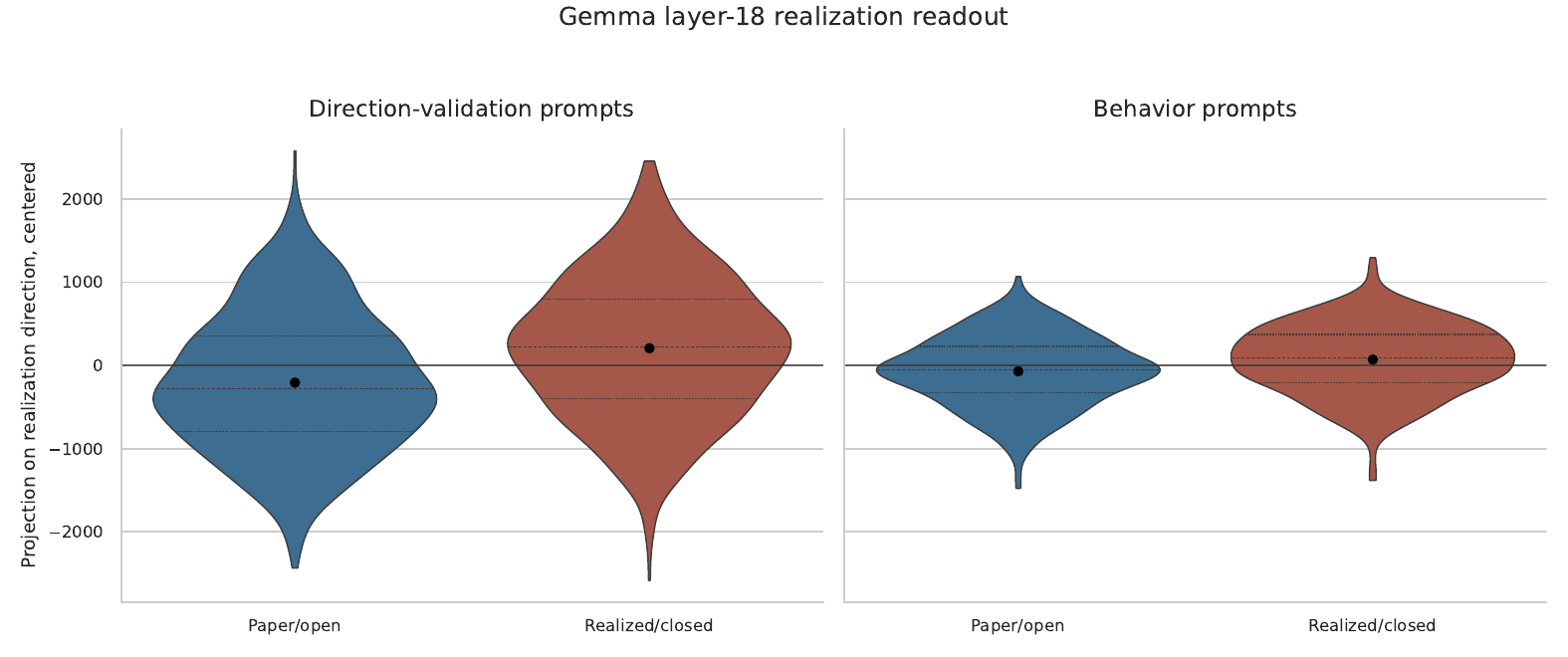}
  \caption{Gemma layer-18 projections on the all-pairs realization direction, centered within split. Realized/closed prompts shift upward relative to paper/open prompts across all logged splits. This direction was built from the full activation set and should be treated as descriptive rather than a strict generalization test.}
  \label{fig:activation-projection-appendix}
\end{figure}

\newpage

\section{Train-only Held-out Readout Results}
\label{app:readout}

\begin{table}[h!]
  \centering
  \small
  \caption{Train-only layer-18 readout by split. Mean projection delta is the average realized/closed minus paper/open projection difference. Correct direction is the percentage of matched pairs where the realized/closed prompt projects higher.}
  \label{tab:heldout-readout-appendix}
  \begin{tabular}{llrrr}
    \toprule
    Dataset & Split & Pairs & Mean projection delta & Correct direction \\
    \midrule
    Original & \texttt{direction\_train} & 756 & 417.36 & 89.2\% \\
    Original & \texttt{direction\_val} & 756 & 413.43 & 91.1\% \\
    Original & \texttt{behavior\_eval} & 324 & 137.19 & 80.6\% \\
    DeepSeek held-out & \texttt{heldout\_readout} & 28 & 443.62 & 92.9\% \\
    DeepSeek held-out & \texttt{heldout\_behavior\_eval} & 12 & 123.08 & 75.0\% \\
    \bottomrule
  \end{tabular}
\end{table}

\newpage

\section{Steering Compliance by Scale and Prompt Source}
\label{app:compliance}

\begin{figure}[h!]
  \centering
  \includegraphics[width=0.92\textwidth]{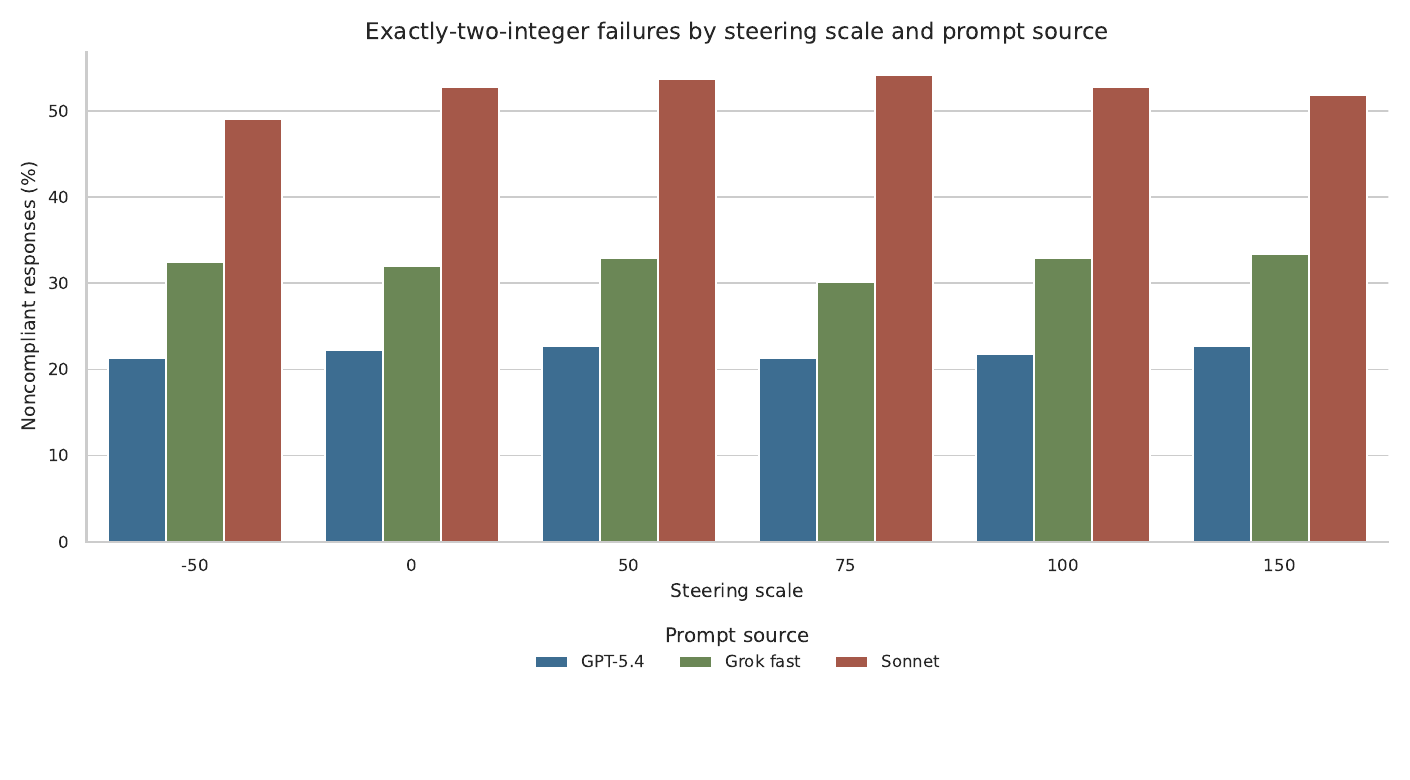}
  \caption{Exactly-two-integer noncompliance by steering scale and prompt source. Sonnet-derived prompts fail at substantially higher rates than GPT-5.4-derived or Grok-fast-derived prompts, suggesting compliance failures reflect an interaction between prompt phrasing and the steered generation process rather than a uniform effect of steering magnitude.}
  \label{fig:compliance-appendix}
\end{figure}

\end{document}